\documentclass[letterpaper, 9pt, conference]{ieeeconf}  

\IEEEoverridecommandlockouts                              

\usepackage{epsfig} 
\usepackage{amsmath} 
\usepackage{amssymb}  
\usepackage{bm}
\usepackage{dsfont}
\usepackage{color}
\usepackage{cite}
\usepackage{colortbl} 
\usepackage{diagbox}
\usepackage[absolute,overlay]{textpos}
\usepackage[linesnumbered,ruled]{algorithm2e}
\usepackage[normalem]{ulem} 
\makeatletter
\let\NAT@parse\undefined
\makeatother
\usepackage{hyperref}
\hypersetup{pdfstartview=FitH,
            colorlinks=true,
            linkcolor=red,
            anchorcolor=blue,
            citecolor=black
            }
\usepackage{float}
\usepackage{booktabs}
\usepackage{arydshln}
\usepackage{bbm}
\usepackage[caption=false,font=footnotesize]{subfig}
\usepackage[dvipsnames]{xcolor}

\usepackage{soul}
\usepackage{multirow}
\usepackage{multicol}
\usepackage{graphicx}
\usepackage{subfloat}
\usepackage{amsmath}
\usepackage{amssymb}
\usepackage{booktabs}
\newcommand{\etal}{\textit{et al}.}
\newcommand{\eg}{\textit{e.g.}}
\newcommand{\ie}{\textit{i.e.}}

\title{\LARGE \bf
Towards Dynamic and Small Objects Refinement for Unsupervised Domain Adaptative Nighttime Semantic Segmentation
}

\author{Jingyi Pan$^{1}$, Sihang Li$^{1}$, Yucheng Chen$^{1}$, Jinjing Zhu$^{1}$ and Lin Wang$^{1,2}{^*}$
\thanks{$^{*}$Corresponding author}
\thanks{$^{1}$J. PAN, S. Li, Y. Chen, J. Zhu are with the AI Thrust, HKUST(GZ), Guangdong, 511458, China. Email: {\tt\small \{jpan305, sli886, ychen208, jzhu706\}@connect.hkust-gz.edu.cn}}%
\thanks{$^{1,2}$L. Wang is with AI/CMA Thrust, HKUST(GZ) and Dept. of CSE, HKUST, Hong Kong SAR, China, Email: {\tt\small linwang@ust.hk}  
}}

\newcommand{\statement}{\noindent\begin{textblock*}{\textwidth}(2cm,0.5cm)\large \centering This work has been submitted to the IEEE for possible publication. Copyright may be transferred without notice, after which this version may no longer be accessible.\end{textblock*}}

\begin{document}

\statement

\maketitle
\thispagestyle{empty}
\pagestyle{empty}

\begin{abstract}

Nighttime semantic segmentation plays a crucial role in practical applications, such as autonomous driving, where it frequently encounters difficulties caused by inadequate illumination conditions and the absence of well-annotated datasets. Moreover, semantic segmentation models trained on daytime datasets often face difficulties in generalizing effectively to nighttime conditions. Unsupervised domain adaptation (UDA) has shown the potential to address the challenges and achieved remarkable results for nighttime semantic segmentation. However, existing methods still face limitations in 1) their reliance on style transfer or relighting models, which struggle to generalize to complex nighttime environments, and 2) their ignorance of dynamic and small objects like vehicles and poles, which are difficult to be directly learned from other domains.
This paper proposes a novel UDA method that refines both label and feature levels for dynamic and small objects for nighttime semantic segmentation. First, we propose a dynamic and small object refinement module to complement the knowledge of dynamic and small objects from the source domain to target the nighttime domain. These dynamic and small objects are normally context-inconsistent in under-exposed conditions. Then, we design a feature prototype alignment module to reduce the domain gap by deploying contrastive learning between features and prototypes of the same class from different domains, while re-weighting the categories of dynamic and small objects.
Extensive experiments on three benchmark datasets demonstrate that our method outperforms prior arts by a large margin for nighttime segmentation. Project page: \url{https://rorisis.github.io/DSRNSS/}.

\end{abstract}

\section{Introduction}
Semantic segmentation is essential for scene understanding of  autonomous driving~\cite{hofmarcher2019visual,siam2018comparative,blum2019fishyscapes, liu2021coinet,malone2022improving, roggiolani2023hierarchical} and robotic systems~\cite{kerkech2020vine, palazzo2020domain, liu2021light, katuwandeniya2021multi}.
Recently, with the development of deep neural networks (DNNs), semantic segmentation has achieved remarkable progress. However, existing methods predominantly target the daytime images~\cite{zou2018unsupervised,poudel2019fast,guizilini2021geometric}. Thus their performance drastically drops in challenging environments,~\eg, nighttime~\cite{sakaridis2019guided,sakaridis2020map,wu2021dannet}, especially to dynamic and small objects. There are many approaches leveraging other modalities to assist the nighttime segmentation, \ie, thermal cameras~\cite{vertens2020heatnet, sun2019rtfnet} and event cameras~\cite{zhang2021issafe, cao2023chasing}, but the bottleneck of single-modal semantic segmentation at nighttime lies in the decreased performance resulting from inadequate illumination and the scarcity of annotated datasets.

\begin{figure}[t!]
\begin{center}
\captionsetup{font=small}
\includegraphics[width=\linewidth]{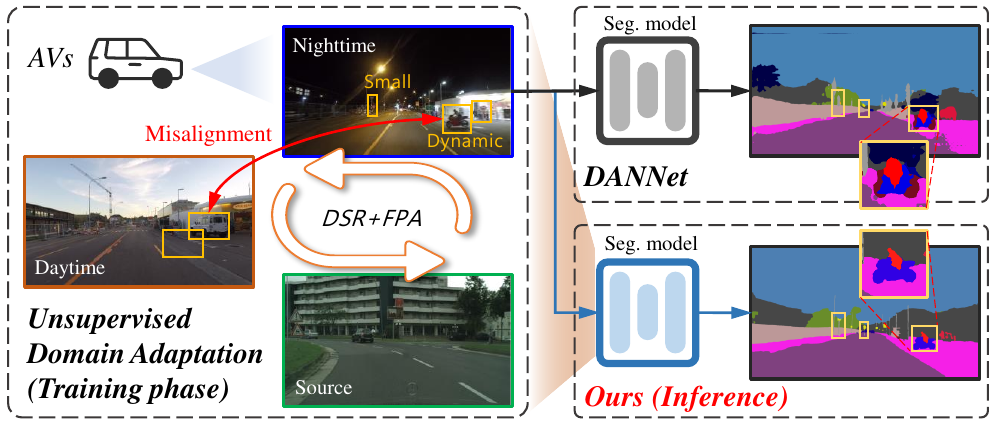}
\end{center}
\vspace{-15pt}
\caption{\textbf{Left:} Our UDA method addresses the difficulty of transferring knowledge about dynamic and small objects from other domains for nighttime segmentation. \textbf{Right:} Compared with DANNet~\cite{wu2021dannet}, our method significantly improves the performance of the dynamic and small objects.}
\label{fig:onecol}
\vspace{-12pt}
\end{figure}

To tackle the problem, unsupervised domain adaptation (UDA) methods have been developed to adapt the model learning unlabeled target domain (\ie, nighttime) images from the model trained with the source domain (\ie, daytime) images. The prevailing methods can be categorized into three types: 1) Some methods,~\eg,~\cite{wulfmeier2017addressing, wulfmeier2018incremental, sakaridis2019guided, sun2019see, sakaridis2020map,gao2022cross}, use the style transfer models, \eg, CycleGAN~\cite{zhu2017unpaired}, to generate daytime or nighttime images, which serve as an intermediate domain to connect the source and target domains. However, these methods are cumbersome as they require multi-stage learning, and the performance can't be guaranteed if style transfer fails.
2) Other methods utilize twilight images with the coarsely aligned day-night image pairs in the target domain, to progressively adapt from the daytime to nighttime domains~\cite{dai2018dark, sakaridis2019guided, sakaridis2020map}. 
3) The rest methods leverage prior GPS knowledge or static loss to reduce the influence of coarsely aligned day-night image pairs, which improves the quality of pseudo labels~\cite{wu2021dannet,yang2021bi,lee2022gps,liu2022improving,shen2023loopda}. However, they pay less attention to \textit{\textbf{the dynamic and small objects}},~\eg, vehicles and poles, in nighttime images, which are particularly challenging to transfer from the daytime domain to the nighttime domain due to their misalignment as shown in Fig.~\ref{fig:onecol}. Also, existing UDA methods employing patch-wise or pixel-wise contrastive learning~\cite{alonso2021semi, liu2021domain} face challenges in \textbf{\textit{acquiring effective semantic contexts within patches or pixels in the under-exposed conditions}}.





\textbf{Motivation:} In this paper, we propose a novel one-stage UDA framework for nighttime semantic segmentation paying more attention to the dynamic and small objects. Specifically,
our framework consists of two key technical modules: dynamic and small objects refinement (\textbf{DSR}) module and feature prototype alignment (\textbf{FPA}) module. Firstly, the DSR module generates a composite mask from the ground truth of the source domain, emphasizing dynamic and small objects, and randomly selected classes. We also introduce a memory bank to store the long-tailed objects across images, which complements these low-occurring categories from different images (Sec.~\ref{sec_3.2}). 
We observe that the mixup operation forms a new domain\textemdash the mixed domain that has a domain shift with the source daytime and target nighttime domain. Therefore, we propose the FPA module to align the source domain with both the mixed and target nighttime domain, aiming to generally reduce the domain gap (Sec.~\ref{sec_3.3}). FPA extracts reliable prototypes from labels in the source domain and pseudo-labels in the mixed domain while pulling the features with the same classes from another domain as positive pairs closer, and pushing other negative pairs away via a contrastive loss. Note that the same operation is conducted in the source and target nighttime domain. We further design an adaptive re-weighting mechanism to improve the attention of some dynamic and small categories in the FPA module. Extensive experiments on the Dark Zurich ~\cite{sakaridis2019guided}, the Nighttime Driving~\cite{dai2018dark}, and ACDC~\cite{sakaridis2021acdc} datasets outperform the state-of-the-art (SoTA) in most categories, particularly in identifying dynamic and small objects. Notably, our approach achieved remarkable mean intersection over union (mIoU) scores of 60.9\% for the 'pole' category, 86.8\% for the 'car' category, and 25.2\% for the 'bus' category, which represents a significant improvement of \textbf{2.9\%},\textbf{ 3.0\% }and \textbf{20.2\%} absolute performance, respectively, over the SoTA method.


In summary, the main contributions of this work are three-fold: (\textbf{I}) We study a crucial problem of focusing more on improving the performance of dynamic and small objects and reducing the domain shifts caused by illumination and style differences.
(\textbf{II}) We propose a novel UDA framework for nighttime semantic segmentation framework by designing the dynamic and small objects refinement (DSR) module and feature prototype alignment (FPA) module. (\textbf{III}) Extensive experiments on the Dark Zurich, Nighttime Driving, and ACDC-night datasets verify that our method achieves a new SoTA performance for nighttime semantic segmentation.


\begin{figure*}[h]
\captionsetup{font=small}
\begin{center}
\includegraphics[width=\linewidth]{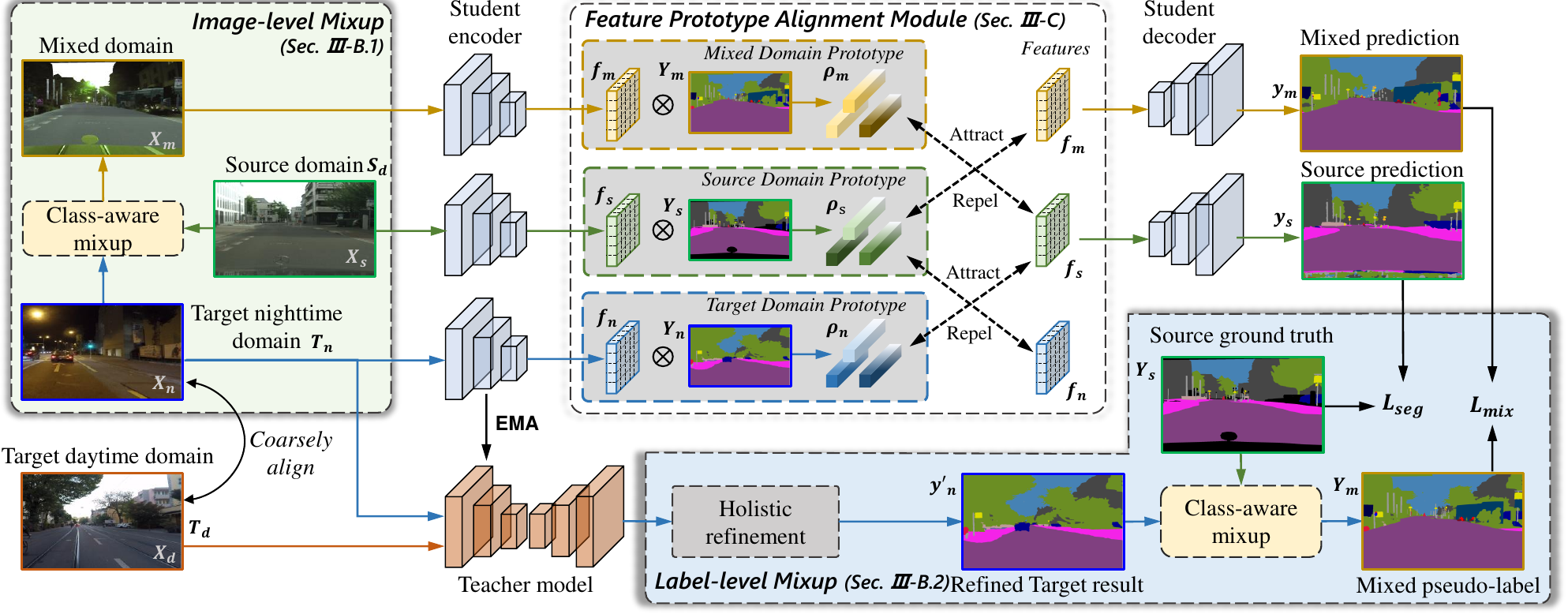}
\end{center}
\vspace{-18pt}
   \caption{Overview of our proposed framework. $F_s$ and $F_t$ denote the student network and teacher network. 1) In the image-level mixup stage, we mix dynamic and small classes in source images into target nighttime images according to the ground truth of source images. 2) In the FPA module, we calculate prototypes ($\rho_m$, $\rho_s, \rho_n$) of each class to regularize the pixel embedding space ($f_s$, $f_m$, $f_n$) with contrastive learning.  3) In the label-level mixup stage, the refined target results obtained from the holistic refinement and source ground truth are mixed in the same classes as the image-level mixup into the mixed pseudo label for focusing the dynamic and small objects refinement.}
\label{framework}
\vspace{-12pt}
\end{figure*}

\vspace{-2pt}
\section{Related work}
\vspace{-3pt}
\subsection{UDA for semantic segmentation}
\vspace{-3pt}
\subsubsection{Daytime semantic segmentation}
Daytime semantic segmentation can be divided into three categories. 
The first type of approach leverages adversarial learning to reduce the domain gap~\cite{hoffman2016fcns, tsai2018learning, liu2021coinet}. 
The second type of method employs style transfer models,~\eg, CycleGAN~\cite{zhu2017unpaired}, to build an intermediate domain to bridge the domain gap\cite{he2021multi,ma2021coarse,isobe2021multi}.
The rest line of the methods impose extra constraints on data distribution to align source and target domains for domain adaptation. \cite{wang2021exploring} utilizes the NCE loss on a vector which is converted from the feature to obtain the global semantic relationship. On the other hand, Liu \etal \cite{liu2021source} bring the distribution closer to each other by BN statistical information.
However, these daytime semantic segmentation models are hard to be applied in nighttime scenarios due to the poor illumination. Thus, we focus on improving the performance of nighttime semantic segmentation, especially for dynamic and small objects.
\subsubsection{Nighttime semantic segmentation} 
\cite{wulfmeier2017addressing, romera2019bridging, sakaridis2019guided, sakaridis2020map, sun2019see, wu2021dannet,yang2021bi} utilize style transfer models to enhance the nighttime appearance to resemble daytime lighting. Among them, DANNet~\cite{wu2021dannet} designs a relighting network to push the intensity distribution of daytime and nighttime images closer. Bi-Mix~\cite{yang2021bi} proposes a bidirectional mixing framework and exploits day-night image pairs to improve the quality of relighting. However, these methods render the segmentation performance to be highly dependent on the style transfer model.
Besides, several works propose to construct an intermediate domain to reduce the domain gap. 
\cite{dai2018dark,sakaridis2020map,sakaridis2019guided} progressively adapt the daytime-trained model to model learning nighttime images, assisted by the intermediate twilight domain. Also, some methods use other sensors to assist nighttime segmentation~\cite{vertens2020heatnet,xia2023cmda}.
\textit{However, these methods ignore the dynamic and small objects, \eg, vehicles and poles, in the domain alignment process, and the capacity to address the inherent domain shifts between datasets caused by illumination and style differences}.
\subsection{Mixup}
\vspace{-2pt}
It is a strategy to augment data to improve the robustness of DNNs. It has been widely used to augment samples between the source and target domains to reduce the domain gap. \cite{mao2019virtual} \cite{xu2020adversarial} \cite{wu2020dual} exploit the mixup ratio combined with randomly sampled values from the beta distribution to achieve domain adaptation. However, mixup samples are inclined to have local ambiguity and artifacts, which result in model confusion. 
Cutmix~\cite{yun2019cutmix} is similar to general mixup while the difference is that it removes pixels and adds removed regions with regions from another image. After that, Walawalkar \etal \cite{walawalkar2020attentive} propose attentive cutmix to enhance the cutmix strategy for better generalization. Classmix~\cite{olsson2021classmix}  is a generalization of cutmix, which utilizes a binary mask to mix randomly sampled images. \textit{Unlike previous mixup strategies, we propose a dynamic and small object refinement strategy, which enables the nighttime domain to enhance these weak categories from the source domain, \ie, dynamic and small objects.}


\section{Method}
\subsection{Overview}
An overview of our proposed framework is shown in Fig.~\ref{framework}. 
The proposed approach involves a labeled source domain $S_d$ and two coarsely aligned, unlabeled target domains $T_d$ and $T_n$. In the source domain, we have pixel-level annotated source data $X_s \in S_d$ and extract feature representations $f_s$ using the student model $F_s$. We then obtain segmentation predictions $y_s$ supervised by cross-entropy loss $\mathcal{L}{sup}$. Furthermore, the target daytime $X_d \in T_d$ and nighttime images $X_n \in T_n$ are unlabeled. We obtain segmentation predictions $y_d$ and $y_n$ using the teacher network $F_t$ and align the large static regions present in both daytime and nighttime target images through holistic refinement~\cite{bruggemann2022refign}. By replacing the aligned regions in nighttime predictions with those of daytime predictions, we obtain a refined nighttime pseudo label $y^{\prime}_{n}$. Our aim is to learn a target model $F^{\prime}_s$ that obtains more knowledge on the dynamic and small objects from labeled source data and unlabeled target daytime domain.

The lack of emphasis on dynamic and small objects in previous works has negatively impacted the nighttime segmentation performance. Moreover, domain shifts caused by illumination and style differences among different domains have further complicated the issue. Thus, we first propose a dynamic and small objects refinement (DSR) module (Sec.~\ref{sec_3.2}) that employs a class-aware mixup function to generate a composite mask from ground truth in the source domain and then adds small and dynamic objects from the source images to nighttime images, which creates a mixed domain. This mixed domain aims to provide complementary supervision for these classes in nighttime images (Sec.~\ref{sec_3.2.1}). To further enhance the effectiveness of our mixup, we obtain more reliable pseudo-labels for nighttime images using holistic refinement. We then use label-level mixup to generate mixed pseudo labels to effectively supervise the mixed images from image-level mixup (Sec.~\ref{sec_3.2.2}). Additionally, we introduce the memory bank mechanism further to complement rare dynamic and small objects across different images. Moreover, we propose the feature prototype alignment (FPA) module among the source domain, mixed domain, and target nighttime domain to decrease inherent domain shifts. We also introduce a re-weighting strategy to pay more attention to the dynamic and small classes (Sec.~\ref{sec_3.3}). Detailed descriptions of these modules are provided below.

\subsection{Dynamic and Small Object Refinement}
\label{sec_3.2}
\subsubsection{Image-level Mixup}
\label{sec_3.2.1}

\begin{figure}[t!]
\captionsetup{font=small}
\begin{center}
\includegraphics[width=1\linewidth]{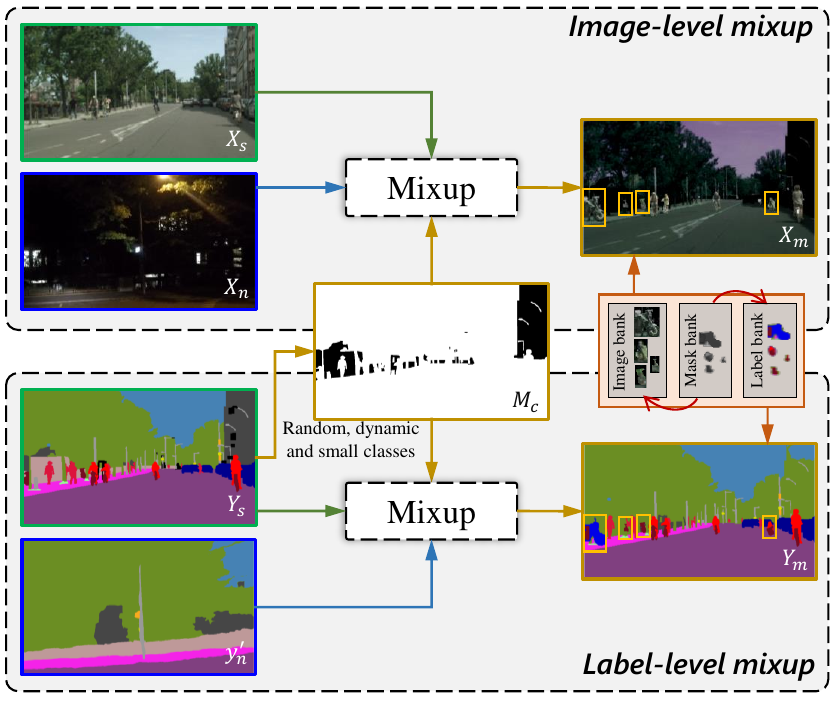}
\end{center}
\vspace{-15pt}
   \caption{An illustration of our DSR module. At the image level, two images are obtained from $S_d$ and $T_n$, respectively. We generate a composite mask from $Y_s$ in the source domain, including randomly selected, dynamic and small classes. This mask is then used to mix the two images while also mixing $Y_s$ and $y^{\prime}_n$. Furthermore, the long-tailed memory bank mechanism introduces a few occurrence regions into the mixed image and label.}
   \vspace{-15pt}
\label{framework_LMRR}
\end{figure}

Owing to the coarse alignment between target nighttime images and their daytime counterparts, there is a lack of direct supervision for dynamic and small objects in nighttime scenes, making it challenging to obtain accurate pseudo labels from the daytime domain. To address these challenges, we first leverage the labels of the source images to mix up the regions of dynamic and small objects from the source domain into the nighttime images for dynamic and small object refinement (See Fig.~\ref{framework_LMRR}). By doing so, we can provide accurate predictions on these previously overlooked objects. 

Based on the labels $Y_s$ of source images, we define the mixed mask as:
{\setlength\abovedisplayskip{2pt}
\setlength\belowdisplayskip{2pt}
\begin{equation}
    M(h,w) = \left\{
    \begin{array}{ll}
        1, & \mbox{if } Y_s(h,w) \in c \\
        0, & \mbox{otherwise } 
    \end{array}
    \right.
\label{compute_mask},
\end{equation}}
where $M$ is a binary mask, and $c$ is the selected class. The parameters $h \in H$ and $w \in W$ represent the height and width of the image, respectively.


We further generate a composite mask, denoted as $M_c$, including the randomly selected class mask $M_r$ and the dynamic and small object mask $M_m$ that contains classes of small and dynamic objects in the source images. 
{\setlength\abovedisplayskip{2pt}
\setlength\belowdisplayskip{2pt}
\begin{equation}
    M_c = M_r \cup M_m.
\label{Mc}
\end{equation}}
Based on the composite mask $M_c$, we augment the nighttime images with complementary objects from the source domain, especially for dynamic and small objects.
For image-level mixup, we utilize the composite mask to mixup corresponding regions of source images and target images as follows:

{\setlength\abovedisplayskip{2pt}
\setlength\belowdisplayskip{2pt}
\begin{equation}
    X_m =  M_c \odot X_n + (1 - M_c) \odot X_n,
\label{compute_xm}
\end{equation}}
here, $\odot$ represents element-wise multiplication.
\subsubsection{Label-level Mixup} 
\label{sec_3.2.2}
To improve supervision within the mixed image domain, it is essential to generate reliable pseudo labels through label-level alignment facilitated by the composite mask, as outlined in Eq.~\ref{Mc}. Thus, we mixup the labels of source and nighttime images corresponding to the image-level mixup as:
{\setlength\abovedisplayskip{2pt}
\setlength\belowdisplayskip{2pt}
\begin{equation}
    Y_m = M_c \odot Y_s + (1 - M_c) \odot y^{\prime}_n
\label{compute_ym},
\end{equation}}
where $y^{\prime}_n$ is the refined pseudo labels of nighttime images generated using target daytime images. 


In addition, addressing the challenge of long-tailed classes, such as buses and bicycles, requires strategies that ensure these classes appear frequently enough to be learned effectively. To this end, we incorporate instances of long-tailed classes from a variety of source images into the target nighttime domain. To facilitate this, we have devised three memory banks: an image bank $B_i$, a mask bank $B_m$, and a label bank $B_l$, which serve to preserve information pertaining to long-tailed classes from different images. These memory banks operate on a first-in-first-out principle, retaining a specified quantity of content at any given time.
Next, we mixup the regions of long-tailed classes from the source domain to the above mixed images ($X_m, Y_m$) as follows:

\vspace{-2pt}
{\setlength\abovedisplayskip{1pt}
\setlength\belowdisplayskip{1pt}
\begin{equation}
\begin{split}
    X^{\prime}_m = B_m \odot B_i + (1 - B_m) \odot X_m,\\
    Y^{\prime}_m = B_m \odot B_l + (1 - B_m) \odot Y_m.
\label{compute_xm2}
\end{split}
\end{equation}}
\vspace{-2pt}

Furthermore, we utilize the student model $F_s$ to make predictions $y^{\prime}_m = F_s(X^{\prime}_m)$ of mixed image $X^{\prime}_m$. 
To ensure consistency between the mixed predictions from $F_s$ and the mixed pseudo labels from $F_t$, we define a mixup loss as follows:
{\setlength\abovedisplayskip{2pt}
\setlength\belowdisplayskip{2pt}
\begin{equation}
    \mathcal{L}_{mix} = -\sum^{C}_{c=0} \sum^{W}_{w=0} \sum^{H}_{h=0} Y^{\prime c,h,w}_{m} log(y^{\prime c,h,w}_{m})
\label{mixuploss}.
\end{equation}}
The weights $\phi^{\prime}_t$ of the teacher model during training are updated at every t step by the student's weights $\phi_t$:
\begin{equation}
\phi^{\prime}_t = \lambda \dot \phi^{\prime}_{t-1} + (1 - \lambda) \dot \phi_t,
\label{ema}
\end{equation}
where $\lambda$ is the EMA decay of the smoothing coefficient, and $\lambda \in [0,1]$.
\subsection{Feature Prototype Alignment}
\label{sec_3.3}
In the context of the DSR module, the mixup operation generates a new domain, referred to as the mixed domain. However, the mixed domain introduces a new domain shift that needs to be addressed for the effective alignment of this domain with the source and target nighttime domains. To overcome this issue, certain constraints must be imposed. In this study, we propose a feature prototype alignment approach to align the mixed domain with the source and target nighttime domain. Note that We also implement alignment of the source domain with the target nighttime domain using a similar approach to effectively reduce the domain gap across the different domains.

To learn domain-invariant features, we aggregate pixel-level features for the same object class across different domains. Firstly, we obtain prototypes of each class by using the ground truth $Y_s$ from the source domain and mixed pseudo label $Y^{\prime}_m$ from the mixed domain:

{\setlength\abovedisplayskip{2pt}
\setlength\belowdisplayskip{2pt}
\small
\begin{equation}
     \rho_s^{c} = \frac{\sum\limits_{h}^{H}\sum\limits_{w}^{W}f_{s}^{h,w}Y_s^{c,h,w}}{\sum\limits_{h}^{H}\sum\limits_{w}^{W}Y_s^{c,h,w}}, 
    \rho_m^{c} = \frac{\sum\limits_{h}^{H}\sum\limits_{w}^{W}f_{m}^{h,w}Y^{\prime c,h,w}_m}{\sum\limits_{h}^{H}\sum\limits_{w}^{W}Y^{\prime c,h,w}_m}   
\label{proto},
\end{equation}}
where $\rho_s^{c}$ and $\rho_m^{c}$ represent the prototypes of the source domain and mixed domain for class $c$, while $f_s$ and $f_m$ denote the source and mixed domain extracted by the student model $F_s$, respectively. We use these prototypes to compute the cross-domain contrastive loss with the features in another domain. Specifically, we select pixel-wise features and the prototype of the same classes between the source domain and mixed domain as positive pairs, while others are set as negative pairs. This helps us to maximize the pixel-prototype similarity within the same classes:
\begin{table*}[ht]
  \centering
  \captionsetup{font=small}
  \caption{Comparison with the state-of-the-art methods and baseline models on the Dark Zurich-test set. The best results are presented in \textbf{bold}.}
  \resizebox{\linewidth}{!}{
  \begin{tabular}{l|cccccccc|ccccccccccc|c}
\toprule
 
  \multirow{2}{*}{Method}& \rotatebox{90}{road} & \rotatebox{90}{sidewallk} & \rotatebox{90}{building} & \rotatebox{90}{wall} & \rotatebox{90}{fence} & \rotatebox{90}{vegetation } & \rotatebox{90}{terrain} & \rotatebox{90}{sky} & \rotatebox{90}{pole} & \rotatebox{90}{light} & \rotatebox{90}{sign}  & \rotatebox{90}{person} & \rotatebox{90}{rider} & \rotatebox{90}{car}  & \rotatebox{90}{truck} & \rotatebox{90}{bus}  & \rotatebox{90}{train} & \rotatebox{90}{motocycle }& \rotatebox{90}{bicycle} & \multirow{2}{*}{mIoU} \\ \cline{2-20}
  & \multicolumn{8}{c|}{Large static objects} & \multicolumn{11}{c|}{Dynamic and small objects}& \\ \hline
RefineNet~\cite{lin2017refinenet}-Cityscapes & 68.8 & 23.2 & 46.8 & 20.8 & 12.6  & 43.1 & 14.3 & 0.3 & 29.8 & 30.4 & 26.9& 36.9 & 49.7 & 63.6 & 6.8 & 0.2 & 24.0 & 33.6 & 9.3 & 28.5\\
GCMA~\cite{sakaridis2019guided} & 81.7 & 46.9 & 58.8 & 22.0 & 20.0  & 64.8 & 31.0 & 32.1 & 41.2 & 40.5 & 41.6 & \textbf{53.5} & 47.5 & 75.5 & 39.2 & 0.0 & 49.6 & 30.7 & 21.0 & 42.0\\
MGCDA~\cite{sakaridis2020map} & 80.3 & 49.3 & 66.2 & 7.8 & 11.0  & 64.1 & 18.0 & 55.8& 41.4 & 38.9 & 39.0 & 52.1 & \textbf{53.5} & 74.7 & 66.0 & 0.0 & 37.5 & 29.1 & 22.7 & 42.5\\
DANNet~\cite{wu2021dannet}-RefineNet &90.0 & 54.0 & 74.8 & 41.0 & 21.1  & 72.0 & 26.2 & 84.0 & 25.0 & 26.8 & 30.2& 47.0 & 33.9 & 68.2 & 19.0 & 0.3 & 66.4 & 38.3 & 23.6 & 44.3\\
Bi-Mix~\cite{yang2021bi} & 89.2 & 59.4 & 75.8 & 41.7 & 19.2  & 70.9 & 30.1 & 81.9 & 39.0 & 31.9 & 31.5& 44.9 & 41.8 & 66.3 & 34.2 & 1.0 & 61.1 & 47.4 & 14.6 & 46.5\\
CCDistill~\cite{gao2022cross} & 89.6 & 58.1 & 70.6 & 36.6 & \textbf{22.5}  & 68.3 & \textbf{33.0} & 80.9 & 33.0 & 27.0 & 30.5& 42.3 & 40.1 & 69.4 & 58.1 & 0.1 & \textbf{72.6}& \textbf{47.7} & 21.3 & 47.5\\ \hline
\textbf{Ours (RefineNet)} & \textbf{93.9} & \textbf{68.2} & \textbf{77.4} & \textbf{32.3} & 11.9  & \textbf{72.9} & 32.4 & \textbf{85.7} & \textbf{50.4} & \textbf{45.4} & \textbf{44.2} & 42.7 & 30.4 & \textbf{75.8} & \textbf{92.9} & \textbf{2.5} & 59.8& 15.6& \textbf{28.4}& \textbf{50.7} \\ \hline\hline
DeepLab-v2~\cite{tsai2018learning}-Cityscapes  & 79.0 & 21.8    & 53.0   & 13.3 & 11.2  & 43.5   & 10.4    & 18.0 & 22.5 & 20.2  & 22.1 & 37.4  & 33.8 & 64.1  & 6.4 & 0.0   & 52.3  & 30.4   & 7.4   & 28.8 \\
DANNet~\cite{wu2021dannet}-DeepLab-v2       & 88.6 & 53.4    & 69.8   & 34.0 & \textbf{20.0}   & \textbf{69.5}   & 32.2   & \textbf{82.3} & 25.0 & 31.4 & 35.9 & 44.2   & 43.7  & 54.1 & 22.0  & 0.1 & 40.9   & 36.0  & 24.1   & 42.5 \\
SePiCo ~\cite{xie2023sepico} & 91.2 & 61.3 & 67.0 & 28.5 & 15.5  & 65.4 & 22.5 & 80.4 & 44.7 & 44.3 & \textbf{41.3} & 41.3 & \textbf{52.4} & 71.2 & 39.3 & 0.0 & 39.6 & 27.5 & 28.8  & 45.4\\\hline
\textbf{Ours (DeepLab-v2)} & \textbf{94.3} & \textbf{68.9} & \textbf{77.4} & \textbf{37.4} & 17.5  & 67.6 & \textbf{35.4} & \textbf{82.3} & \textbf{46.6} & \textbf{45.6} & 41.2 & \textbf{53.7} & 47.7 & \textbf{79.6} & \textbf{71.3} & \textbf{6.7} & \textbf{71.7} & \textbf{40.8} & \textbf{34.5} & \textbf{53.7}\\ \hline\hline 
HRDA~\cite{hoyer2022hrda}-Cityscapes  & 90.4 & 56.3 & 72.0 & 39.5 & 19.5  & 59.3 & 29.1 & 70.5 & 57.8 & 52.7 & 43.1& 60.0 & \textbf{58.6} & 84.0 & 75.5 & 11.2 & 90.5 & \textbf{51.6} & 40.9 & 55.9\\
Refign~\cite{bruggemann2022refign}-HRDA & \textbf{95.4} & \textbf{76.1}  & \textbf{86.9}  & \textbf{53.7} & \textbf{37.1}   & 79.4  & \textbf{42.0}  & 91.3 & 58.0 & \textbf{57.2}  & \textbf{63.7}& \textbf{63.8}  & 56.6 & 83.8 & \textbf{85.6} & 5.0 & 91.7 & 47.4 & 39.5 & 63.9\\ \hline 
\textbf{Ours (HRDA)} & 95.2 &	75.1 &	85.0 &	52.0 &	36.2  &	\textbf{80.2} &	38.8 	& \textbf{91.8} &	\textbf{60.9} &	56.9 &	63.5 &	62.5 &	57.3 &	\textbf{86.8} &	\textbf{85.6} &	\textbf{25.2} &	\textbf{92.1} &	\textbf{51.6} &	\textbf{42.1} &\textbf{65.2} \\
\bottomrule
\end{tabular}}
 \label{tab:darkzurich}
 \vspace{-5pt}
\end{table*}

\begin{figure*}[h]
\captionsetup{font=small}
\begin{center}
\includegraphics[width=\linewidth]{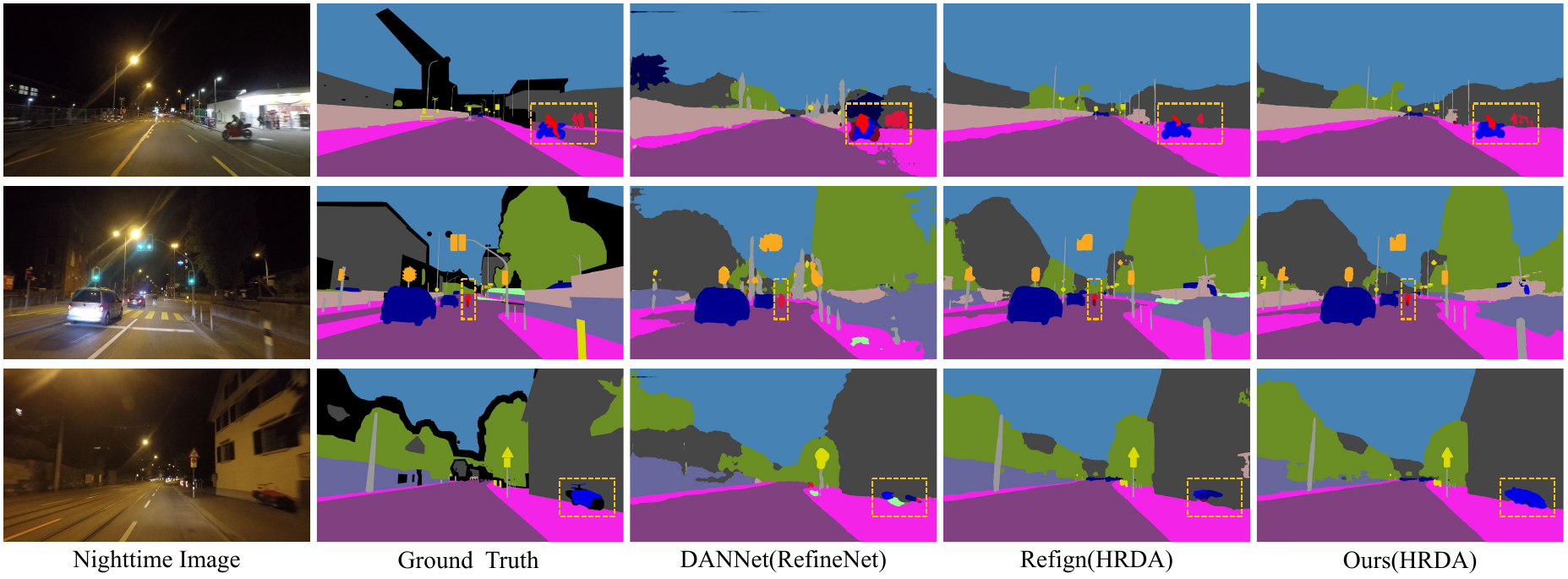}
\end{center}
\vspace{-10pt}
   \caption{\textbf{Qualitative} comparison between our approach and the state-of-the-art methods on the \textbf{Dark Zurich-val set}. Better viewed when zoomed in.}
\label{quality}
\vspace{-8pt}
\end{figure*}
{\setlength\abovedisplayskip{2pt}
\setlength\belowdisplayskip{2pt}
\begin{equation}
    S_{m \rightarrow s}^{c} = (s(f_m^{h,w},\rho_{s}^{c,h,w})/\tau)\cdot W^{c},
\label{s_con2}
\end{equation}}
where $s(\cdot,\cdot)$ denotes cosine similarity, and $\tau$ is a temperature parameter, while $W^{c}$ is the similarity weight of class $c$. We can calculate the contrastive loss as follows.
{\setlength\abovedisplayskip{2pt}
\setlength\belowdisplayskip{2pt}
\begin{equation}
    \mathcal{L}_{m \rightarrow s} = -\sum_{c}^{C}\sum_{h}^{H}\sum_{w}^{W}y^{\prime c,h,w}_{n}log\frac{exp(S_{m \rightarrow s}^{c})}{\sum_{c}exp(S_{m \rightarrow s}^{c})},
\label{source_con2}
\end{equation}}
here, $\mathcal{L}_{m \rightarrow s}$ represents the contrastive loss that measures the distance between the source prototypes and mixup features. Additionally, the prototypes of the mixed domain can effectively capture positive pairs belonging to the same class in the source domain while simultaneously repelling negative pairs.
{\setlength\abovedisplayskip{2pt}
\setlength\belowdisplayskip{2pt}
\begin{equation}
\begin{split}
    S_{s \rightarrow m}^{c} = (s(f_s^{h,w},\rho_{m}^{c,h,w})/\tau)\cdot W^{c},\\
    \mathcal{L}_{s \rightarrow m} = -\sum_{c}^{C}\sum_{h}^{H}\sum_{w}^{W}Y_{s}^{c,h,w}log\frac{exp(S_{s \rightarrow m}^{c})}{\sum_{c}exp(S_{s \rightarrow m}^{c})}.
\end{split}
\label{s_con2}
\end{equation}}
To address the issue of class imbalance, we propose an adaptive re-weighting algorithm specifically for cases where the overlapping regions contain dynamic and small objects. By assigning weights to these prototypes, we can adjust the corresponding pixel-wise features accordingly.
We set the weights of cosine similarity as:
{\setlength\abovedisplayskip{2pt}
\setlength\belowdisplayskip{2pt}
\begin{equation}
    W^{c} = \left\{
    \begin{array}{ll}
        1, & \mbox{if } c \in C_o \\
        s+1/s & \mbox{if } c \in C_l\\
        0, & \mbox{otherwise } 
    \end{array}
    \right.
\label{reweight}
\end{equation}}
where $W$ denotes the weight of the contrastive loss, and $s$ represents the number of overlapping classes. The set $C_o$ contains the overlapping classes while not including the long-tailed classes, whereas $C_l$ comprises the long-tailed classes that are also part of the overlapping classes. The total prototype contrastive loss is:
 \begin{equation}
     \mathcal{L}_{proto} = \mathcal{L}_{n \rightarrow s} + \mathcal{L}_{s \rightarrow n} + \mathcal{L}_{m \rightarrow s} + \mathcal{L}_{s \rightarrow m}.
\label{protoloss}
 \end{equation}
\subsection{Objective functions}
In addition to the loss functions described in Sec.~\ref{sec_3.2} and Sec.~\ref{sec_3.3}, we also utilize cross-entropy loss to supervise our segmentation model with labeled source images.
{\setlength\abovedisplayskip{2pt}
\setlength\belowdisplayskip{2pt}
\begin{equation}
\mathcal{L}_{sup} = -\sum^{H}_{h=0}\sum^{W}_{w=0}\sum^{C}_{c=0}\mathds{1}(Y_s^{c,h,w})log(y_{s}^{c,h,w}),
\label{eq}
\end{equation}}
where $\mathds{1}(\cdot)$ is the one-hot encoding operation.

The objective of the entire network is formulated as:
{\setlength\abovedisplayskip{2pt}
\setlength\belowdisplayskip{2pt}
\begin{equation}
\mathcal{L} = \mathcal{L}_{sup} + \alpha{\mathcal{L}_{mix}} + \beta{\mathcal{L}_{proto}}.  
\label{eq}
\end{equation}}
where $\alpha$ and $\beta$ are hyper-parameters to balance three terms. In this work, we set the values as follows: $\alpha=1.0$, $\beta=0.1$.

\section{Experiments}

\subsection{Datasets}
For all experiments, we use the mean of category-wise Intersection-over-union (mIoU) as the evaluation metric. The following datasets are utilized for model training and performance evaluation:

\noindent\textbf{Cityscapes}~\cite{cordts2016cityscapes} This dataset contains 5,000 frames with pixel-level annotations of 19 categories from street scenes in 50 cities. The images have a resolution of 2,048 x 1,024 pixels. It consists of 2,975 training images, 500 validation images, and 1,525 testing images. 

\noindent\textbf{Dark Zurich}~\cite{sakaridis2019guided} The Dark Zurich dataset includes 3,041 daytime images, 2,920 twilight images, and 2617 nighttime images captured with coarse aligned in three conditions. We utilize 2,416 day-night image pairs for training, 151 for testing, and 50 for validation. 

The input for our source domain comes from Cityscapes~\cite{cordts2016cityscapes}, while the input of our target domain consists of daytime and nighttime images from Dark Zurich~\cite{sakaridis2019guided}.
Additionally, we evaluate our nighttime segmentation model on Nighttime Driving test set~\cite{dai2018dark} and ACDC-night val set~\cite{sakaridis2021acdc}.

\subsection{Implementation details}
We implement our method by using Pytorch-lightning on single Nvidia RTX A6000 GPU. We choose HRDA~\cite{hoyer2022hrda}, DeepLabv2~\cite{tsai2018learning} and RefineNet~\cite{lin2017refinenet} as our based methods. Both our models are trained by the AdamW~\cite{loshchilov2017decoupled} optimizer with a weight decay of $1 \times 10^{-2}$, and the initial learning rate is set as $6 \times 10^{-4}$, and linear learning rate warmup with warmup ratio $10^{-6}$. Then the learning rate is decreased with the poly learning rate policy with a power of 0.9. The batch size is set to 2. The total training iterations are set to 40k. An EMA factor is $\alpha=0.999$. At the inference time, there is no change to be introduced to the final model $F_s$. The training resolution follows the used UDA methods (half resolution for DeepLabv2, and RefineNet and full resolution for HRDA).

\begin{table*}[t!]
    \centering
    \footnotesize
    \setlength{\tabcolsep}{0.01\linewidth}
    \captionsetup{font=footnotesize}
    \caption{Comparison with state-of-the-art methods on Nighttime Driving test sets and Dark Zurich-val sets. The best results are presented in \textbf{bold}. ‘D’, means DeepLab-v2\cite{tsai2018learning}, 'R' means RefineNet\cite{lin2017refinenet}, and 'T' means HRDA\cite{hoyer2022hrda}. }
    \resizebox{0.6\linewidth}{!}{
    \begin{tabular}{lccccc}
        \toprule
        \multirow{2}{*}{Method} & \multirow{2}{*}{Backbone}&\multicolumn{4}{c}{mIoU}\\\cline{3-6}
        \multicolumn{2}{c}{} & Nighttime Driving & Dark Zurich-val& ACDC-night  \\ \hline
        RefineNet~\cite{lin2017refinenet}-Cityscapes& R & 32.8 & 15.2& 20.3 \\
        DeepLab-v2~\cite{tsai2018learning}-Cityscapes& D & 25.4 & 12.1& 16.3 \\
        GCMA~\cite{sakaridis2019guided}& R & 45.6 & 26.7& - \\
        MGCDA~\cite{sakaridis2020map}& R & 49.4 & 26.1& - \\
        DANNet~\cite{wu2021dannet}-RefineNet& R & 42.4 & 30.0& 37.0 \\
        CCDistill~\cite{gao2022cross}& R & 46.2 & -& 37.3 \\ 
        Refign~\cite{bruggemann2022refign}-HRDA& T & 58.0 & 47.5& 53.2 \\\hline
        \textbf{Ours (RefineNet)}& R &50.8 & 34.5 &40.3  \\
        \textbf{Ours (DeepLabv2)}& D & 52.4 & 37.7 &43.6  \\
        \textbf{Ours (HRDA)}& T & \textbf{58.4} & \textbf{48.5}& \textbf{53.8}\\
        \bottomrule
    \end{tabular}}
    
    \label{tab:nighttimedriving}
    \vspace{-15pt}
\end{table*}

\subsection{Comparison with state-of-the-art methods}

\noindent\textbf{Comparison on Dark Zurich.} We compare our proposed method with some existing state-of-the-art methods, incuding GCMA~\cite{sakaridis2019guided}, MGCDA~\cite{sakaridis2020map}, DANNet~\cite{wu2021dannet}, Bi-Mix~\cite{yang2021bi}, CCDistill~\cite{gao2022cross}, and several other domain adaptation methods~\cite{hoyer2022hrda, xie2023sepico, bruggemann2022refign}. In order to maintain a fair comparison, we employ a consistent backbone when comparing each method and present the results of corresponding source-only models. The outcomes are reported in Tab.~\ref{tab:darkzurich}. 

As seen in Tab.~\ref{tab:darkzurich}, our method delivers an increase in mIoU of approximately \textbf{1.3\%}, surpassing the state-of-the-art method Refign (HRDA)~\cite{bruggemann2022refign}. Furthermore, our method demonstrates significant performance improvements over other methods in several categories, particularly in misaligned regions such as vegetation, sky, pole, car, truck, bus, train, motorcycle, and bicycle, where bus and bicycle are categories with relatively fewer occurrences. While scores for other categories differ slightly from the best scores, we achieve competitive performance in each backbone. Our methods exhibit significant improvements over other techniques, particularly in detecting dynamic and small objects. Although there is a degree of randomness in the scores across various categories, the results of our method with three different backbones demonstrate clear advantages in identifying the majority of dynamic and small objects.

Furthermore, we present qualitative results on Dark Zurich-val in Fig.~\ref{quality}, which corroborate our quantitative results. The qualitative results reveal that our method can accurately predict intricate details of small and dynamic objects, such as the predictions of a person's legs in the first and second rows, and the motorcycle's details in the third row.

\begin{figure}[t]
\captionsetup{font=small}
\begin{center}
\includegraphics[width=\linewidth]{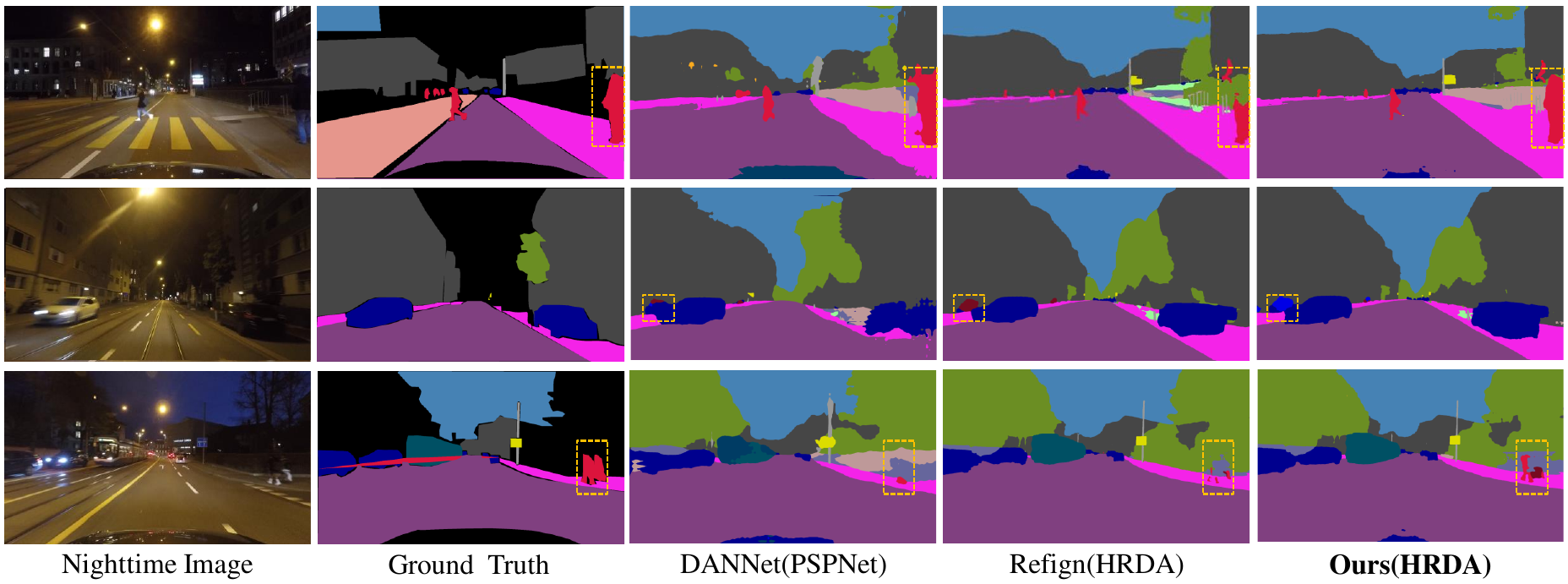}
\end{center}
\vspace{-10pt}
   \caption{\textbf{Qualitative} comparison between our approach and some existing state-of-the-art methods on the \textbf{Nighttime driving test set}. Better viewed when zoomed in.}
\label{r1}
\vspace{-10pt}
\end{figure}

\begin{figure}[t]
\captionsetup{font=small}
\begin{center}
\includegraphics[width=\linewidth]{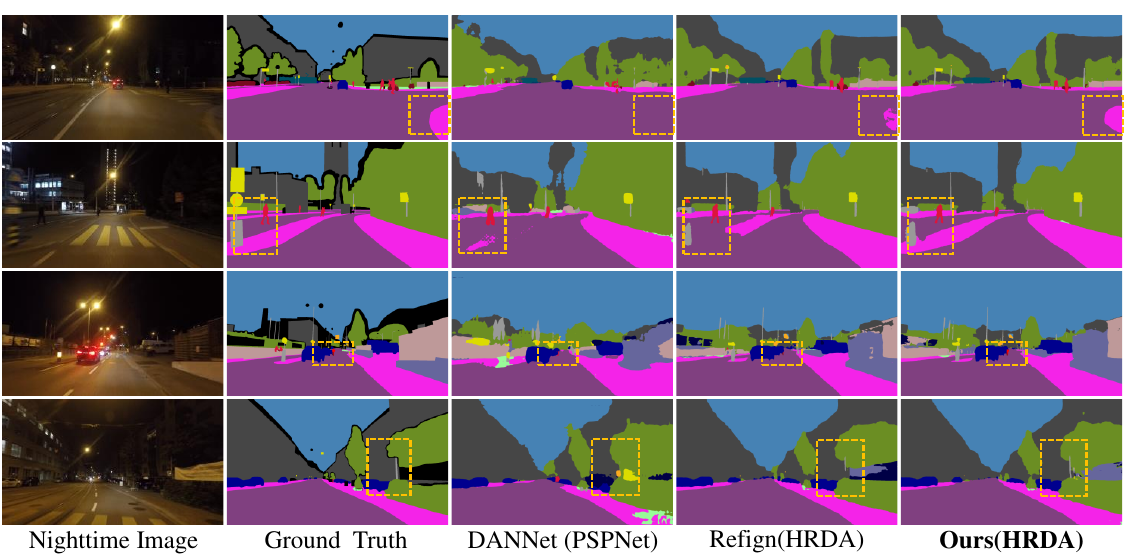}
\end{center}
\vspace{-10pt}
   \caption{\textbf{Qualitative} comparison between our approach and the state-of-the-art methods on the \textbf{ACDC-night val set}. Better viewed when zoomed in.}
\label{r2}
\vspace{-18pt}
\end{figure}

\noindent\textbf{Comparison on other Datasets.} To confirm the generality of our approach, we present the performance of our method and compare results with previous methods on the Nighttime driving test set, and ACDC-night val set in Tab. \ref{tab:nighttimedriving}. Our method with HRDA achieves a \textbf{0.4\%} and \textbf{0.6\%} increase in mIoU over the best score in the Nighttime driving test set and ACDC-night val set, respectively, while the performance of our method outperforms other methods with the same backbone. In Fig~\ref{r1} and Fig.~\ref{r2}, we provide qualitative results of Nightime driving test set and ACDC-night val set that clearly demonstrate the enhanced accuracy of our method in predicting the details of dynamic and small objects. 
It is worth noting that the Nighttime driving test set has coarse annotation. Despite this limitation, our method still achieves significant performance improvements with different backbones. 

\subsection{Ablation study}
In this section, we train several model variants for 40k iterations and test them on Dark Zurich-test due to their fine-grained pixel-wise annotation and completed categories.

\noindent\textbf{Effectiveness of each component.} we conduct extensive experiments to validate the effectiveness of different components of our method with the same settings and hyperparameters. The results are presented in Tab.~\ref{tab:component}. Adding the composite mask $M_c$  for input and dynamic and small objects refinement improved performance to 63.2 mIoU. Incorporating the long-tailed bank further increased performance by 0.5\%. The prototype loss $L_{proto}$ and re-weighting strategy successfully addressed domain shifts and improved alignment between domains.

\begin{table}[t]
    \centering
    \footnotesize
    \setlength{\tabcolsep}{0.005\linewidth}
    \captionsetup{font=small}
    \caption{Ablation study of each component.}
    \vspace{-5pt}
    \resizebox{\linewidth}{!}{
    \begin{tabular}{cccccc}
        \toprule
        Baseline~\cite{hoyer2022hrda} & DSR(w/o long-tailed bank) & DSR(full) & FPA(w/o re-weighting) & FPA(full) & mIoU\\ \hline 
        \checkmark & \multicolumn{4}{c}{} & 55.9\\
        \checkmark & \checkmark & \multicolumn{3}{c}{} & 63.2\\
        \checkmark & \checkmark & \checkmark & \multicolumn{2}{c}{} & 63.7\\
        \checkmark & \checkmark & \checkmark & \checkmark & \multicolumn{1}{c}{} & 64.8\\
        \checkmark & \checkmark & \checkmark & \checkmark & \checkmark & 65.2\\
        \bottomrule
    \end{tabular}}
    \label{tab:component}
    \vspace{-8pt}
\end{table}

\noindent\textbf{Effectiveness of the DSR selected classes.}
To demonstrate the effectiveness of our DSR module, 
we select different regions in our method and train them with the same implementation for a fair comparison. The results are presented in Tab.~\ref{tab:mixup}. 
Our method, combining randomly selected classes and misaligned regions, achieved a +5.2\% mIoU improvement for the baseline and +2.7\% mIoU improvement for the randomly selected classes method. This validates the effectiveness of our domain adaptive nighttime semantic segmentation approach. 
\begin{table}[t!]
    \centering
    \captionsetup{font=small}
    \caption{Ablation results of selected mixup classes in DSR module.}
    \vspace{-5pt}
    \resizebox{0.8\linewidth}{!}{
    \begin{tabular}{lccccc}
        \toprule
        Method &\multicolumn{3}{c}{}& mIoU & $\Delta$ mIoU ($\%$)\\ \hline 
        Baseline + FPA & \multicolumn{3}{c}{} & 60.0 & -\\
        + random classes~\cite{olsson2021classmix}& \multicolumn{3}{c}{} & 62.5 & +2.5\\
        + random classes + small classes & \multicolumn{3}{c}{} & 63.7 & +3.7\\
        + random classes + dynamic classes & \multicolumn{3}{c}{} & 64.1 & +4.1\\ \hline
        Ours & \multicolumn{3}{c}{} & \textbf{65.2} & \textbf{+5.2}\\
        \bottomrule
    \end{tabular}}
    
      \label{tab:mixup}
      \vspace{-8pt}
\end{table}

\noindent\textbf{Effectiveness of the loss weight.}
We perform a sensitivity analysis on our method's hyperparameters $\alpha$ and $\beta$ to evaluate their impact on performance, as presented in Tab.~\ref{tab:loss_weights}. By varying the values of $\alpha$ and $\beta$, we measure their effect on the mIoU. The results indicate that the mIoU remains relatively stable when $\beta$ remains fixed and $\alpha$ increases significantly.
Based on these results, we select the initial weights of $\alpha = 1$ and $\beta = 0.1$. 


\begin{table}[!t]
  \centering
  \begin{minipage}[t]{0.5\linewidth}
    \centering
    \captionsetup{font=small}
    \small
    \caption{Ablation results on loss weights. $\alpha$ and $\beta$.}
    \vspace{-8pt}
    \resizebox{1\linewidth}{!}{
    \begin{tabular}{c|cccc}
    \toprule
    \diagbox{$\beta$}{$\alpha$}& 0.75  & 1.00 & 1.25  & 1.50  \\\hline
    0.075 & 62.94 & 63.66          & 63.41 & 63.93 \\
    0.100 & 63.99 & \textbf{65.18} & 64.77 & 64.78 \\
    0.125 & 62.37 & 64.91          & 63.21 & 62.65\\
    \bottomrule
    \end{tabular}}
    \vspace{-10pt}
    \label{tab:loss_weights}
  \end{minipage}%
  \vspace{2pt}
  \begin{minipage}[t]{0.4\linewidth}
    \centering
    \small
    \caption{Performance on Cityscapes-val.}
    \vspace{-5pt}
    \label{tab:cityscapes}
    \resizebox{\linewidth}{!}{
    \begin{tabular}{lc}
    \hline
     Method & MIoU\\ \hline
     DAFormer~\cite{hoyer2022daformer} &68.3\\
     HRDA~\cite{hoyer2022hrda} & 73.8\\
     ours (HRDA) & \textbf{81.7}\\ 
    \hline
    \end{tabular}}    
    \vspace{-10pt}
  \end{minipage}%
\end{table}

\vspace{-2pt}

\subsection{Discussion}


\subsubsection{Refinement of dynamic and small objects}
\begin{figure}[!t]
\captionsetup{font=small}
\begin{center}
\includegraphics[width=\linewidth]{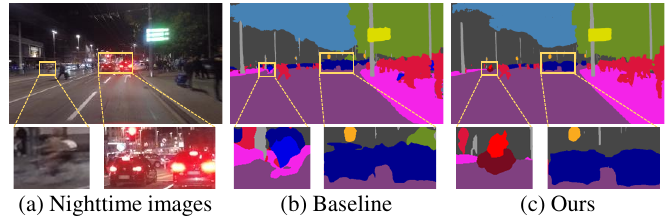}
\end{center}
\vspace{-10pt}
   \caption{Detailed visualization of our method for refining dynamic and small objects with Dark Zurich-test set~\cite{sakaridis2019guided}.}
\label{r5}
\vspace{-15pt}
\end{figure}

As depicted in Fig.~\ref{r5}, despite the complexity of the environment and the presence of numerous dynamic and small objects, such as people, cars, poles, lights, and bicycles, our segmentation results demonstrate clearer edges and more accurate categorization compared to the baseline. This is the case even in conditions of blur and extremely complex light.

\subsubsection{All-day performance}
Our method is designed to prioritize nighttime image segmentation while ensuring that it remains highly effective for daytime images when compared to established UDA baseline models in Cityscapes-val dataset~\cite{cordts2016cityscapes}, as shown in Tab.~\ref{tab:cityscapes}. Overall, our experimental results demonstrate the ability of our method to segment all-day images effectively.

\section{Conclusion}
In this paper, we proposed a novel UDA framework for nighttime segmentation by focusing on dynamic and small object refinements and inherent domain shifts. The DSR module combines the random selection of classes with dynamic and small objects while introducing a long-tailed bank to store long-tailed objects to augment the dynamic and small data, which are hard to segment due to poor illumination.
The FPA module further constrains the mixup process through feature-level prototype alignment to reduce the domain shift in different domains while improving the weights of dynamic and small objects during the alignment process.
Experimental results demonstrated the effectiveness of our method.



{\small
\bibliographystyle{IEEEtran}
\bibliography{main}
}


\end{document}